\documentclass{article}

\usepackage[preprint]{nips_2018}

\usepackage[utf8]{inputenc} 
\usepackage[T1]{fontenc}    
\usepackage{hyperref}       
\usepackage{url}            
\usepackage{booktabs}       
\usepackage{amsfonts}       
\usepackage{nicefrac}       
\usepackage{microtype}      
\usepackage{caption}
\usepackage{listings}
\usepackage{graphicx}
\usepackage[section]{placeins}
\title{Low Resource Text Classification with ULMFit and Backtranslation}

\author{
  Sam Shleifer \\ 
  Stanford University \\ 
 \texttt{shleifer [at] stanford.edu} \\ 
}

\begin{document}

\maketitle

\begin{abstract}
In computer vision, virtually every state of the art deep learning system is trained with data augmentation. In text classification, however, data augmentation is less widely practiced because it must be performed before training and risks introducing label noise. 
We augment the IMDB movie reviews dataset with examples generated by two families of techniques: random token perturbations introduced by \citet{wei} and backtranslation -- translating to a second language then back to English. 

In low resource environments, backtranslation generates significant improvement on top of the state-of-the-art ULMFit model.\footnote{\citet{ulm}} A ULMFit model pretrained on wikitext103 and then finetuned on only 50 IMDB examples and 500 synthetic examples generated by backtranslation achieves 80.6\% accuracy, an 8.1\% improvement over the augmentation-free baseline with only 9 minutes of additional training time. Random token perturbations do not yield any improvements but incur equivalent computational cost.

The benefits of training with backtranslated examples decreases with the size of the available training data. On the full dataset, neither augmentation technique improves upon ULMFit's state of the art performance. We address this by using backtranslations as a form of test time augmentation as well as ensembling ULMFit with other models, and achieve small improvements.
\end{abstract}

\section{Introduction}
Text Classification is an important problem for many applications, including spam detection, emergency response, and legal document discovery, but is frequently limited by the cost of acquiring large labeled datasets. For this reason, it is important to develop systems that perform adequately in "low resource" settings, where few labeled examples are available. \footnote{\cite{ulm}} Transfer learning from language models trained on larger corpora of text is a leap forward,  allowing models to be initialized with significant understanding of language, but there is still room for improvement. In the IMDB task we study, where the model predicts whether a movie review is positive or negative, it needs roughly 500 example reviews to reach 85\% accuracy and remains below human performance when given access to the full dataset of 25,000 examples.

\section{Related Work}
\subsection{Data Augmentation}
 Data augmentation addresses data scarcity by creating synthetic examples that are generated from, but not identical to, the original document. In computer vision, data augmentation is part of nearly every SOTA model. Images are randomly cropped, brightened, or dimmed during training, in an effort to allow the model to see more diverse examples and learn invariance to changes in input that don't effect its target. In NLP, augmentation is less widely used. \citet{sennrich} and \citet{edunov} show significant improvements in Neural Machine Translation BLEU score by training their system on its own outputs, in a technique they named \textit{backtranslation}. In text classification, there is little academic research on the relevant extension to their work -- generating synthetic training examples by using an external system to translate a training document to another language and then back to the original language. We refer to this analagous technique by the same name: backtranslation.

\begin{table}[!htbp]

\begin{tabular}{@{}ll@{}}
\toprule
Operation & Sentence                                                                     \\ \midrule
None      & A sad human comedy played out on the back roads of life.           \\
          &                                                                              \\
BT (Spanish)       & A sad human comedy that develops in the secondary roads of life. \\
          &                                                                              \\
BT (Bengali)       & A sad man played the street behind comedy life. \\
          &                                                                              \\
Synonmym Replace$^\dagger$        & A \underline{lamentable} human comedy played out on the \underline{backward} road of life. \\
          &                                                                              \\
Random Insert$^\dagger$         & A sad human comedy played out on \underline{funniness} the back roads of life. \\
          &                                                                              \\
Random Swap$^\dagger$         & A sad human comedy played out on \underline{roads back the} of life.           \\
          &                                                                              \\
Random Delete$^\dagger$         & A sad human\underline{ }out on the \underline{ } roads of life.                              \\ \bottomrule
\end{tabular}
\caption{BT stands for backtranslation. $^\dagger$ Token Perturbation techniques from Wei and Zhou, [2019]}
\label{wei_table}
\end{table}

The different data augmentation techniques we experimented with are presented in Table \ref{wei_table}. For token level transformations we use code and ideas from \citet{wei}, who randomly perturb 10 \% of tokens with one of the four $^\dagger$ transformations, before training an LSTM and CNN models from scratch.  Over 6 text classification tasks, they report 0.8\% improvement over an 87.8\% average for full datasets and 3.0\% improvement over a 76.9\% average baseline when training on 500 original examples. The only pretrained component of the authors' model is word embeddings, and our results suggest that token level perturbations do not offer improvements on top of ULMFit, a stronger baseline built on a pretrained language model.

\subsection{Architecture: ULMFit}
We use \citet{ulm}'s ULMFit architecture for binary text classfication.
ULMFit consists of three phases: 
\begin{enumerate}
\item Train a 3 layer AWD-LSTM\footnote{\citet{awd}} language model on \citet{merity2016pointer}'s wikitext103 data. We download the result of this step from \url{ http://files.fast.ai/models/wt103/}.
\item Fine-tune the language model on the classification dataset (IMDB in our case). Save its encoder. This "in-domain finetuning" step is one of ULMFit's most important contributions.
\item Classification training with CrossEntropy loss:
	\begin{enumerate}
	\item  Let $H_{T-70\ldots T}$ represent the encoder's hidden state for the last 70 words (roughly as many as fit into 16 GB GPU RAM) before $T$, the end of the document.
	\item  concatenate $[h_{T} , maxpool(H), meanpool(H)]$ and feed through a 3 layer fully connected network.
	\end{enumerate}
\end{enumerate}

ULMfit also introduces a number of useful tricks to speed up finetuning, including discriminative learning rates (lower for layers closer to the input), and gradual unfreezing of layers during training (freeze the pretrained encoder at the beginning of stage 3).

Most importantly, ULMFit is the strongest baseline for IMDB. When the authors train separate models on forwards and backwards reviews, using pretrained forwards and backwards language models, then average the predictions, they achieve 4.6$\%$ test error. Without ensembling, their forward model achieves 5.2\% error. We present somewhat worse metrics when we rerun the authors' published code in Table \ref{baselines_table}. ULMFit is implemented inside the fastai repository. \footnote{\url{https://github.com/fastai/fastai/tree/master/courses/dl2/imdb_scripts}}

\section{Approach}

We attempt to use each of the following techniques to improve binary sentiment classfication on IMDB \footnote{\cite{imdb}} movie reviews: (1)  Backtranslation
(2) Token level transformations following \citet{wei}. 
(3) Virtual Adversarial Augmentation following \citet{miyato2016virtual}. (4) Test Time Augmentation, and (5) Ensembling.  We run experiments with methods (1) and (2) on different sizes of input data in Section \ref{res_section}, and discuss results for methods 3-5, which were only attempted on the full dataset, in Section \ref{full_section}. In both sections, the relevant baseline is ULMFit trained on the equivalent data size, and is included in the results.

The \textbf{IMDB Dataset} consists of 100,000 movie reviews. 25,000 are labeled training data, half positive and half negative, 25,000 are labeled test data, and 50,000 do not have labels. In 
Section \ref{res_section}, we do not use the unsupervised examples for language model finetuning, but in Section \ref{full_section} we do. 

431,000 backtranslations were generated using 14\footnote{For 4 languages we only scraped training backtranslations. For another 4 languages we only scraped test backtranslations. For 6 languages we scraped both train and test.} different languages with a Google Translate API offered by the textblob python package. The synthetic examples, along with python code to generate more for arbitrary text, can be found at \url{https://github.com/sshleifer/text-augmentation}.


\section{Low Resource Results}  
\label{res_section}
\begin{figure}[!htbp]
\begin{center}
\includegraphics[scale=0.7]{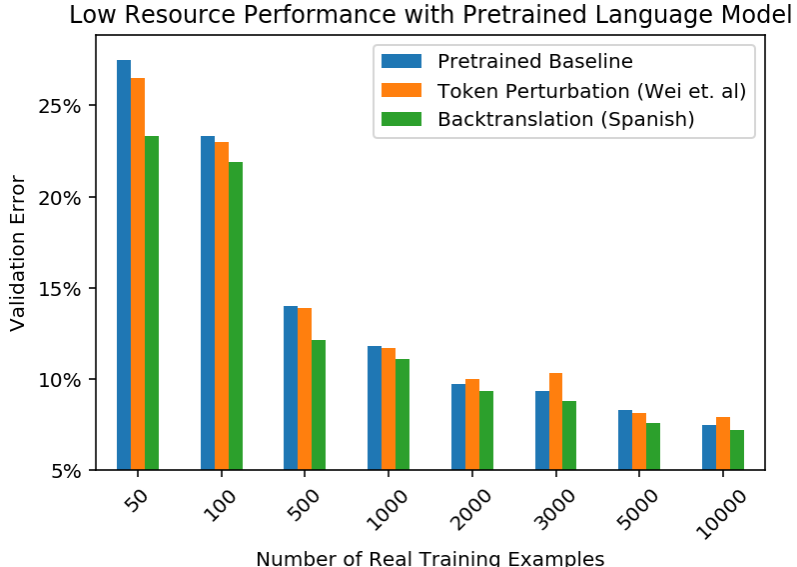}
\caption{
This chart shows ULMFit classification performance of three different methods in low resource settings. The Backtranslation (Spanish) and Token Insertion methods generate one synthetic example for each available training example. Points are the median of 3 separate runs. The improvement generated by backtranslation is the distance between the Blue and Green bars, and ranges from 4.2\% for 50 examples to 0.4\% for 10,000 examples.
}
\label{low_pt}
\end{center}
\end{figure}

Finetuning allows ULMFit to perform well with very few labeled examples, and synthetic examples generated by backtranslation can provide additional gains. Figure \ref{low_pt} shows ULMFit performance with access to different amounts of training data, and suggests that back translation generates larger improvements when used on lower resource models. Synthetic examples are less valuable than original examples, we hypothesize, and so once the model has seen 10,000 real examples and achieved 92.5\% accuracy, 10,000 more synthetic examples generate only .3\% improvement. These results also suggest that it easier to improve a weaker model, but we cannot easily disentagle these overlapping hypotheses.

\begin{figure}[!htbp]
\begin{center}
\includegraphics[scale=0.7]{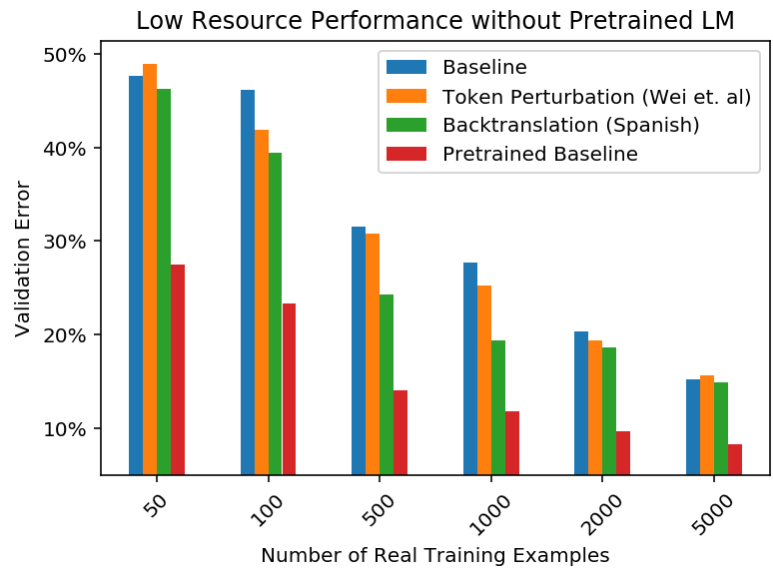}
\caption{Language models were trained from scratch using the limited data, instead of initialized with pretrained wikitext103 weights and then finetuned. ULMFit's Stage 3, classification, was unchanged. The Backtranslation and Token Perturbation methods generate one synthetic example for each available training example. Bars represent the median of 3 separate runs of ULMFit}
\label{low_scratch}
\end{center}
\end{figure}

Figure \ref{low_scratch} shows significantly worse performance if our model is not given access to the pretrained language model. The results suggest that although token perturbation produces larger gains on top of this weaker model, they are still not as large as the improvements generated by backtranslation. Even if we generate 2,4, or 8 Token Perturbation examples for every original example, backtranslation still yields larger gains.

Most importantly, Figure \ref{low_scratch} shows that the gains from pretraining dwarf the gains from either augmentation technique, and persist as we increase data quantity. Practitioners should prioritize using pretrained representations before they consider backtranslation.

\begin{table}[!htbp]
\begin{center}

\begin{tabular}{@{}lll@{}}
\toprule
Languages                & Error@N=50 & Error @N=1000 \\ \midrule
None                     & 0.275            & 0.118               \\
10 Languages                    & \textbf{0.194}            & 0.114               \\
Spanish                  & 0.233            & 0.111               \\
Spanish, French          & 0.225            & \textbf{0.109}               \\
Spanish, French, Bengali & 0.228            & 0.111               \\ 
Bengali & 0.241            & 0.113               \\ \bottomrule
\end{tabular}
\caption{The first column represents the languages through which we generate backtranslations. Points are the median of 3 runs. $\dagger$ Spanish, French, German, Afrikaans,  Russian, Czech, Estonian, Haitian Creole, Bengali. }
\label{donger}
\end{center}
\end{table}

A natural question that emerges from these results is whether it is possible to use more backtranslations to achieve larger gains. The results in Table \ref{donger} show that we observe an additional 4\% gain by adding 500 backtranslated examples from 10 languages generated to the original 50. The second column suggests that so synthetic examples are not needed in the 1,000 example setting, where two backtranslated examples per original example was the best performing configuration, but just one extra example per original performs almost identically. We did not experiment with using backtranslated examples generated from a mix of languages, but would not be surprised if that generated further small improvements. 

\section {Full Dataset Experiments}

\label{full_section}

 In our experiments, training on synthetic examples stopped yielding any improvements once the model had access to more than 15,000 training examples, or 60\% of the dataset. This motivated us to test whether using the backtranslations as a form of test time augmentation might help the model. At the risk of straying off topic, we discuss these results briefly. 

\subsection {Baselines}

\begin{table}[!htbp]
\centering
\begin{tabular}{@{}lll@{}}
\toprule
Method                            & Reported Test Error & Replicated Test Error  \\ \midrule
(A) ULMFit FWD \citet{ulm}                         & 5.30\%   & 5.32\% \\
(B) ULMFit BWD                        &          & 7.38\% \\
(C) A + B                 & \textbf{4.60}\%   & 5.14\% \\
(D) iVAT \citet{sato} & 5.66\%   & 6.24\% \\
(E$^*$)  (A) + TTA        &          & 4.97\% \\
(F$^*$)  (C) + TTA + (D)   &          & \textbf{4.73}\% \\ \bottomrule
\end{tabular}
\caption{$^*$ denotes ensembles created for this project, with weights generated from validation data. The Replicated Test Error column reflects the result when we run the authors published code without modification.}
\label{baselines_table}
\end{table}
Both low-resource and full dataset experiments use ULMFit trained on the same data, with the hyperparemeters from the paper, as a baseline. Although we could not match the authors' reported metrics on the full dataset by running the published code \footnote{And reducing the learning rate for ULMFit backward, as suggested by Sebastian Ruder in an email.} , we came reasonably close, as shown in table \ref{baselines_table}. Our baseline results for low-resource experiments also appear to be very similar to ULMFit paper's Figure 3, which charts their results in low-resource experiments.

\subsection{Test Time Augmentation and Ensembling}

In test time augmentation (TTA), we collect the model's predictions for both real and synthetic examples and combine them in someway, in our case a weighted average optimized on the validation set. We generated predictions for backtranslations through 7 different languages, but low resource languages' like bengali were given 0 weight. As shown in \ref{baselines_table}, TTA generated small improvements (roughly 0.17\%, or 42 extra correct predictions) in the "full data" setting.

Experiments that tried to add Virtual Adversarial Training (VAT), which is explained briefly in the appendix, to ULMFit did not work well. Memory increased threefold, which required a commensurate reduction in batch size. When we ran \citet{sato}'s system that uses a more interpretable variant of VAT, however, and then averaged its predictions with our ULMFit predictions, it yielded another small increase in accuracy. We attribute the diverging results of using VAT in our model and their model to our suboptimal implementation of virtual adversarial loss as well as the extra depth of ULMFit's encoder, which makes computing the required second forward and backward pass more expensive.

\begin{figure}[!htbp]
\begin{center}
\includegraphics[scale=0.7]{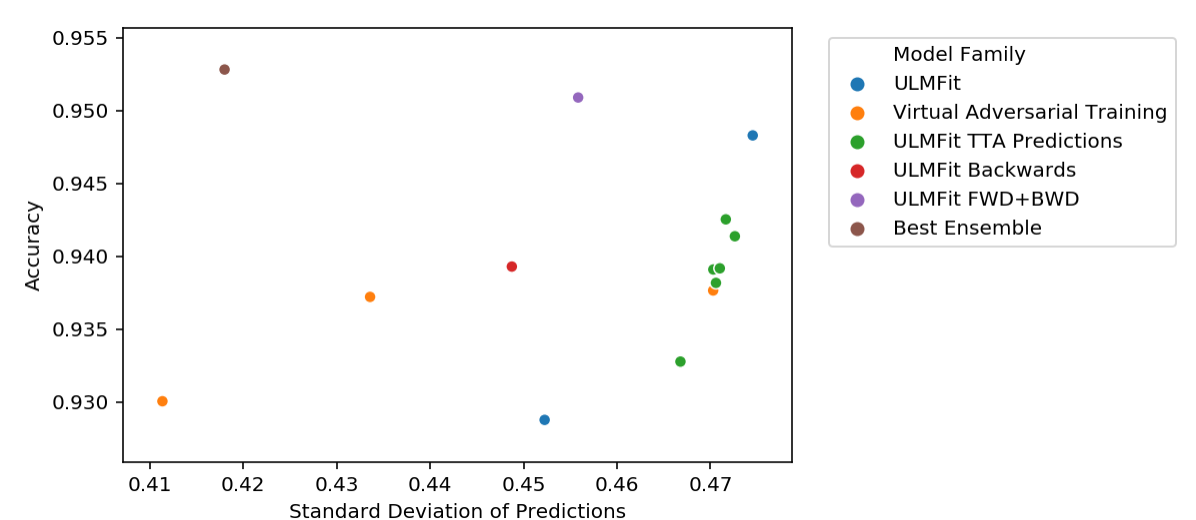}
\caption{Standard deviation of different models' predictions vs. Test accuracy. All models were trained on the full dataset. Green "ULMFit TTA predictions" points refer to the ULMFit forward model's predictions for test set examples backtranslated through another language. The lower green dots tend to represent lower resource languages, with Bengali the least accurate. The lowest blue dot is a ULMFit model whose classifier is trained for only one epoch, with only the last layer unfrozen.}
\label{model_vars}
\end{center}
\end{figure}

\textbf{Overconfident Predictions}: Model predictions tend to be far from 50\%. Figure \ref{model_vars} shows that ensembling multiple models helps reduce variance, but it remains stubbornly high. At the end of training the forward model, over 96\% of train predictions are outside of the 10-90\% range, and 89.8\% are outside of that range for validation data. After ensembling, that statistic falls to 82\%. Figure \ref{model_vars} also suggests that although they are not as accurate, \citet{sato}'s models' predictions (3 orange dots) have lower variance, and a weaker correlation between accuracy and variance, than those of the ULMFit family members. For the ULMFit Forward models (green and blue dots), accuracy and variance of predictions increase in lockstep.

\section{Analysis}

\textbf{ULMFit overweights the last sentence, with good reason.} Reviews average 227 words in length, and many of these words are dedicated to summarizing the plot of the movie. The observation that so much of many reviews is irrelevant for sentiment analysis motivated an experiment to test whether, even after ULMFit's concat pooling, which passes meanpooled and maxpooled representations of all encoder hidden states to the classifier head, ULMFit "overweights" the sentiment of the end of the movie review and underweights the beginning. The regression coefficients shown in Table \ref{coef} suggest that ULMFit does overweight the last sentence, but with good reason -- the last sentence's sentiment is much more predictive of the full document's sentiment than other sentences.  The hypothesis that ULMFit is forgetting important parts of the input is further undermined by the observation that including these same sentence level statistics as features in the ensemble provided no benefit.

\begin{table}[!htbp]
\begin{center}
\begin{tabular}{@{}lll@{}}
\toprule
Feature              & True Label & Prediction \\ \midrule
Avg. Sentence                 & 0.6        & 0.63       \\
Last Sent        & 0.21       & 0.24       \\
First Sent       & 0.03       & 0.07       \\
Max                  & 0          & 0.02       \\
Min                  & 0          & 0.01       \\
Len                  & -0.04      & 0.04       \\
$Model Accuracy$ &  92\%       & 95\%      \\ \bottomrule
\end{tabular}
\caption{Logistic Regression Coefficients estimating the effect of a 1 standard deviation increase in a summary statistic computed over the distribution of sentence level sentiments. One feature, for example, is the average sentiment of sentences in a given review. The experiment design is described further in the appendix.}
\label{coef}
\end{center}
\end{table}

\textbf{Numeracy} In 0.7\% of examples, reviewers end their reviews with a numeric rating of the movie like "Rating: 6/10". The model performs no better on examples with numeric ratings, even though we hypothesize that any review with Rating 5/10 or greater would be labeled positive. ULMFit predicts strong sentiment for these rating strings, but learns an inflection point between 6 and 7. For integer $0 \leq x \leq 6$, the model predicts the sentiment of "Rating x/10" as 0.00, completely negative. For "Rating 6.5/10", the model predicts 38\% chance of positive sentiment, and for 7,8,9 and 10/10 the model jumps to 100\%. 

\textbf{Representative Errors}
Two types of errors appear frequently in ULMFit's output: examples that appear to be mislabeled or completely sarcastic and reviewers disparaging or praising an entity that is NOT the subject of the review. We provide a short example of each type of error, with commentary.

\textbf{Mislabeled or Sarcastic Example:} \textit{"Masterpiece. Carrot Top blows the screen away. Never has one movie captured the essence of the human spirit quite like "Chairman of the Board." 10/10. don\'t miss this instant classic."}

There are about 198 examples in the training dataset where the model's prediction is off by more than 99\%, and when I handlabeled 10 of them I agreed with the model 7 times. An experiment where we remove these suspiciously labeled examples from the training data before training the classifier did not impact metrics. 

\textbf{Misunderstood Reference Example:} \textit{I still enjoyed it even though I hated Othello in high school.} 

Here, the model predicts 100\% negative. Many of the model's errors involve reviews where the author speaks very negatively (positively) about another piece of art to contrast it with a movie they love (hate). The model is not given the title of the movie that is being reviewed, and that information cannot always be inferred from reading the review, making it difficult to track which entity is being disparaged or praised. Named entities are often not the subject of the review, the subject is instead only referred to as "the movie". In this example, however, we can be pretty confident that "it" refers to the movie under review.

\section{Conclusion}
Our work experiments with backtranslation and token perturbation as data augmentation strategies for NLP. We find that backtranslation yields significant improvements over ULMFit in low resource environments, but that these gains disappear when ULMFit is given access to the full dataset. We do not observe gains for token perturbation techniques. 
In the full dataset task, ensembling ULMFit predictions on backtranslated examples with other models' predictions yields small improvements.

\subsection{Future Work}
\textbf{Mixed Backtranslation Strategies} Many interesting experiments were not completed due to lack of time and personnel. One simple extension to this work would be to use backtranslations generated from a mixture of languages. The current set of experiments generate a synthetic example for every combination of training example and language being used for back translation, which means that using an additional language creates an extra $N$ training examples. We suspect that it might be possible to get the same accuracy improvements with fewer synthetic examples by, for example, generating synthetic examples through French for half of the training examples and through Spanish for the other half.

\textbf{Reduce overfitting caused by BT examples} We hypothesize that the backtranslation experiments fail on larger data sizes because of overfitting. Since we do not change ULMFit's learning rate schedule or dropout, the model sees twice as many examples per epoch. In an experiment where ULMFit was trained on the full dataset and the full dataset backtranslated into Spanish, it reached .08 cross entropy train loss before the classifier was fully unfrozen, and achieved no improvement in validation accuracy from that point onward. In an identical experiment without backtranslated examples, however, the train loss was 0.2 at the same point in training, and validation loss increased another 2.2\% over the final 14 epochs of the schedule. The models finished with roughly equal accuracy, but BT trained models might be able to surpass the baseline if they learned a little bit more slowly.

\textbf{Add TTA predictions of other models to the ensemble} Making predictions with the forwards model on backtranslated test examples, and adding those predictions to the ensemble yielded a small increase in accuracy, roughly 0.17\%. It is possible that doing the same with the ULMFit backwards model, or other models, might further improve performance.

\textbf{BERT} Pretrained representations published since ULMFit, like BERT\footnote{\citet{devlin2018bert}}, perform well on many NLP tasks, and it would be interesting to compare them to ULMFit in the low resource and full dataset settings.

\section{Appendix}

\subsection{Virtual Adversarial Training}

In adversarial training, first proposed for NLP by \citet{miyato2016virtual},  the authors select an adversarial perturbation $r$ to the embedding output that minimizes the negative log likelihood of the true class, given $x + r$, subject to the constraint that the norm of $r$ is not too large. Virtual Adversarial training adds another component to that loss function that penalizes the KL-Divergence between the logits of the original model and perturbed model. Whereas adversarial training tries to add a small perturbation to make the model wrong, VAT tries to make the models prediction change in any direction, and was the state of the art model for IMDB classfication before ULMFit. In our final ensemble, we used \citet{sato}'s chainer implementation because it offered a pretrained language model and showed even better results than the original paper.

\subsection{Sentence level Sentiment Regressions}
The regressions in Table \ref{coef} try to predict two targets: the document's label and the model's prediction, using summary stats of the distribution of sentence-level predictions as inputs.  More specifically we fit 
$$target = W\cdot [X_{-1}, X[0], avg(X), max(X), min(X), len(X))]$$, 

where $X$ is a vector whose elements are the model predicted sentiment of each sentiment. We use L1 regularization to get force irrelevant features coefficients to zero. 

\bibliographystyle{unsrtnat}
\bibliography{224_milestone}

\end{document}